\documentclass[]{elsarticle}

\usepackage{setspace}
\usepackage{caption}
\usepackage{graphicx}
\usepackage{subcaption} 
\graphicspath{{fig/}}

\usepackage[T1]{fontenc}

\raggedright
\usepackage{indentfirst}
\usepackage{amsmath}
\usepackage{amssymb}  
\usepackage{txfonts}  
\usepackage{bm}    
\usepackage{hyperref}  
\usepackage{cleveref} 
\usepackage{comment}  
\usepackage{float}  
\usepackage[shortlabels]{enumitem}  
\usepackage{arydshln} 
\usepackage{ulem}  
\usepackage{xcolor} 
\usepackage{booktabs}

\begin{document}

\begin{frontmatter}


\title{Learning Uncertainty with Artificial Neural Networks for Predictive Process Monitoring}

\author{Hans Weytjens}\ead{hans.weytjens@kuleuven.be}\corref{cor1}
\author{Jochen De Weerdt}\ead{jochen.deweerdt@kuleuven.be}
\address{Research Centre for Information Systems Engineering (LIRIS), Leuven, Belgium}


\cortext[cor1]{Corresponding author}


\begin{abstract}
The inability of artificial neural networks to assess the uncertainty of their predictions is an impediment to their widespread use. We distinguish two types of learnable uncertainty: model uncertainty due to a lack of training data and noise-induced observational uncertainty. Bayesian neural networks use solid mathematical foundations to learn the model uncertainties of their predictions. The observational uncertainty can be calculated by adding one layer to these networks and augmenting their loss functions. Our contribution is to apply these uncertainty concepts to predictive process monitoring tasks to train uncertainty-based models to predict the remaining time and outcomes. Our experiments show that uncertainty estimates allow more and less accurate predictions to be differentiated and confidence intervals to be constructed in both regression and classification tasks. These conclusions remain true even in early stages of running processes. Moreover, the deployed techniques are fast and produce more accurate predictions. The learned uncertainty could increase users' confidence in their process prediction systems, promote better cooperation between humans and these systems, and enable earlier implementations with smaller datasets.
\end{abstract}


\end{frontmatter}

%
\section{Introduction}\label{sec:intro}

The goal of process mining is to discover, validate and improve processes by analyzing the time-stamped event logs of information systems. Processes can be driven by humans and/or machines and are omnipresent in business \cite{survey}, manufacturing \cite{manufacturing}, logistics \cite{logistics}, healthcare \cite{healthcare}, and the Internet of Things (IoT) \cite{iot}. Predictive process monitoring is a sub-domain of process mining that involves predicting upcoming events, process outcomes and remaining execution times. Predictive process monitoring has benefited immensely by recent advances in machine learning. Artificial neural networks (NNs) have been established as a state-of-the-art technique in the field.

Despite their power and versatility, however, the adoption of these NNs has been slow in practice. Their black-box nature understandably deters many users. Furthermore, NNs cannot judge the uncertainty of their predictions, reducing user confidence. NNs always make a prediction, even for samples from parts of the domain that were not in the training set or data with significant observational noise. The resulting errors could be expensive or even catastrophic. Therefore, uncertainty awareness is critical.

Various uncertainty quantification techniques have been developed to address this issue. Recent research \cite{UQ, hueller, IEEE, galthesis, uncertainty, galDrop} established strong theoretical foundations for integrating probabilistic theory into neural networks with Bayesian neural networks (BNNs) and proposed methods for practical implementations. BNNs can learn the model, or epistemic, uncertainty caused by a lack of training data. The theoretical framework also includes a mechanism for quantifying the aleatoric uncertainty caused by noisy data. Importantly, the theory also suggests that using these techniques increases the accuracy of the original predictions (point estimates). Unsurprisingly, BNNs have been used in practice in an increasing number of domains and applications \cite{uncertainty, speech, med, astro, bmw, BDLOB}.

To the best of our knowledge, however, no research has been conducted on uncertainty quantification in predictive process monitoring. This paper addresses this research gap. We used neural networks to develop uncertainty-based models for predicting the remaining time and outcomes, evaluating the quality of the uncertainty estimates and measuring the increase in accuracy of the point estimates of the remaining time and outcomes predicted by the theory. Furthermore, we investigated the training and inference speeds and uncertainty applications in predictive process monitoring. We used five publicly available datasets that included event logs of business processes of existing enterprises.
Overall, the quality of the point estimates could be evaluated according to the corresponding uncertainty estimate.
The use of the uncertainty techniques increases the accuracy of the estimates. These results hold for both regression (remaining time prediction) and classification (outcome prediction) tasks and are valid from the beginning of the process when only few events have happened. The uncertainty techniques moderately slow the training and inference times of the NN models.

The remainder of the paper is structured as follows. In Section~\ref{sec:def}, we discuss related work, the contributions of our work, and the motivation for the methods used. In Section~\ref{sec:theory}, we explain the theoretical foundations of our uncertainty quantification techniques. Based on this knowledge, we formulate our research questions in more detail. In Section~\ref{sec: eval}, we introduce the experimental setup, and in Section~\ref{sec:results}, we present our results. We highlight the uncertainty-generated added value for predictive process monitoring applications in Section~\ref{sec:working}. Finally, a summary of our findings and suggestions for future research are discussed in Section~\ref{conclusions}.

\section{Related work}\label{sec:def}
\subsection{Predictive process monitoring}\label{subsec:ppm}
A process is a series of activities that lead to a specific result. A process is similar to, but distinct from, a time series, which is a sequence of data points from measurements taken at uniform time intervals. Process mining involves the study of processes in which humans or technologies such as machines or robots act as agents. Process mining does not include natural processes (e.g., biological and geological processes). Predictive process monitoring is a sub-domain of process mining that involves the prediction of remaining times, process outcomes and future events. The studied processes are neither one-off nor periodic.
Instead, the processes are recurrent and usually occur hundreds or thousands of times, with different degrees of variety.
The datasets include time-stamped logs of such processes, which are referred to as \textit{cases}. These cases consist of \textit{events}. A number of \textit{features} at the event level provide additional details about the events or case. At least one of these features is a \textit{target} feature. In remaining time prediction problems, the target describes the amount of time remaining until the case is completed. In outcome prediction problems, various target features representing the outcomes of cases are possible. Ongoing, incomplete cases are referred to as \textit{prefixes}, with the \textit{prefix length} representing the number of completed events in the prefix. Our task is to train a learner on a training dataset of events with target features and other features that are organized in prefixes, with the objective of predicting the targets of unknown prefixes.\\

Research on remaining time predictions in process mining began with nonparametric regressions \cite{dongen} in 2008. Some years later, \cite{aalst} proposed an annotated transition system. Next, conventional machine learning techniques, including support vector regression and naive Bayes, began to be used in the field \cite{dataaware}. For the first time, attributes other than the activity name and time were included in models. In 2017, \cite{awarelstm, tax} used the long short-term memory model (LSTM), a type of NN. NNs permit automatic feature engineering, eliminating the need for the error-prone, domain-knowledge-based manual feature engineering of conventional machine learning techniques. In 2017, a comparison of remaining time prediction methods \cite{survey} showed that ``the most accurate results were obtained using an LSTM.'' Research on process outcome prediction, which is a classification problem, was similar. For example, \cite{hinkka} explored the use of recurrent neural networks, \cite{cnnlstm} found that convolutional neural networks (CNNs) were faster than LSTMs while providing comparable results, and \cite{kratsch} concluded that deep learning approaches outperformed traditional machine learning techniques in predicting process outcomes. In practice, it is crucial to acquire predictive insights early in running processes. Therefore, most of the aforementioned papers addressed early predictions (\cite{dongen, tax, survey, hinkka, cnnlstm, kratsch}). The superiority of the deep learning methods motivated us to focus on deep NNs in this study.\\

\subsection{Uncertainty}
Nearly all machine learning models always make predictions, even for samples outside the trained domain and samples for which the training data contain very noisy targets. In most cases, these models do not know the uncertainties of their predictions, which can lead to problematic and even catastrophic decisions in practice. This lack of awareness severely reduces the usefulness of prediction techniques, limiting their application. Thus, knowledge on the uncertainty of predictions would be extremely beneficial.

In 1996, \cite{hora} distinguished between epistemic and aleatoric uncertainty, a distinction that is still commonly used in machine learning \cite{hueller, UQ, IEEE, uncertainty}. Epistemic, or model, uncertainty is caused by a lack of data, whereas aleatoric uncertainty is caused by noisy data. \cite{hueller} presents a good overview of methods for learning both types of uncertainty, either combined or separately, as well as a possible taxonomy. Set-based methods \cite{mitchell} such as version learning provide insights into epistemic uncertainty but do not to quantify it. While most machine learning techniques are incapable of directly producing uncertainty estimates, linear regression \cite{linreg} and support vector machine \cite{svm} models do yield uncertainty estimates. Biased and uncalibrated uncertainty estimates can also be obtained based on individual estimates in ensemble techniques such as bagging (for example, random forests \cite{rf}) and boosting. Unfortunately, none of the aforementioned techniques are capable of distinguishing epistemic and aleatoric uncertainty. Gaussian processes \cite{gp} can distinguish these types of uncertainty but do not scale well. Recently, probabilistic, Bayesian neural networks (BNNs) have been emphasized \cite{galthesis}. The quantification of epistemic and aleatoric uncertainty \cite{uncertainty} in both classification and regression tasks relies on solid theoretical foundations. These foundations (which are explained in detail in Section~\ref{sec:theory}) allow for simple BNN implementations, such as NNs with dropout \cite{galDrop}, and in-model learning of the aleatoric uncertainty with more accurate point estimates through loss attenuation \cite{uncertainty}. BNNs are robust against adversarial attacks \cite{robust} with out-of-domain samples, whereas loss attenuation increases model robustness against in-domain outliers with deviating targets (Eq.~\eqref{eq:hetero} in Subsection~\ref{subsec:has}). A detailed discussion of robustness is beyond the scope of this paper. For the above reasons, and because NNs are flexible, universal, state-of-the-art approximators for predictive process monitoring (Subsection~\ref{subsec:ppm}), we focus on NNs in this study.

An overview of uncertainty quantification as it relates to deep learning and various applications is provided in \cite{UQ, IEEE}. Because image processing, computer vision and natural language processing (NLP) are important fields within deep learning, some of the first uncertainty applications appeared here. Following in the footsteps of neural networks, uncertainty has been investigated in domains such as astrophysics \cite{astro}, finance \cite{BDLOB}, chemistry \cite{chem}, medical science \cite{med}, ecology \cite{canopy}, sales \cite{sales}, and manufacturing \cite{bmw}. To the best of our knowledge, however, no research on the application of uncertainty techniques in predictive process monitoring has been conducted.

\subsection{Our contributions}
Our contributions are to apply deep learning uncertainty concepts to predictive process monitoring and to develop uncertainty-based models for regression (remaining time) and classification (outcome) tasks. As a result, we provide an important new application domain for uncertainty quantification techniques while extending existing predictive process monitoring work and increasing its practical use. Our experiments demonstrate that the obtained uncertainty estimates can be used to distinguish between more and less accurate remaining time and outcome predictions, allowing us to rank the predictions and, if necessary, retain only those that meet certain uncertainty thresholds. We also show how confidence intervals can be built based on these uncertainty quantifications. The findings hold even in early process stages for both regression and classification tasks. Moreover, the deployed techniques are fast and, as predicted by the theory, deliver more accurate remaining time and outcome predictions. The learned uncertainty can be used to increase users' confidence in process prediction systems, foster better cooperation between humans and the models and enable faster implementations with smaller datasets.

It should be noted that this paper builds upon our previous work \cite{uncertainty_WW}. This work includes four major extensions and differences. First, the validity of the results is supported by more datasets (five instead of three). Second, the datasets are now unbiased, and there is no leakage between the training and test datasets (see Subsection~\ref{subsec:preprocessing}). Third, we complement the previously researched regression (remaining time) problem with a classification (outcome prediction) problem. Fourth, we present an early prediction analysis.

\section{Estimating uncertainty}\label{sec:theory}
\begin{figure}
\centering
\begin{subfigure}{0.23\textwidth}
    \includegraphics[width=\textwidth]{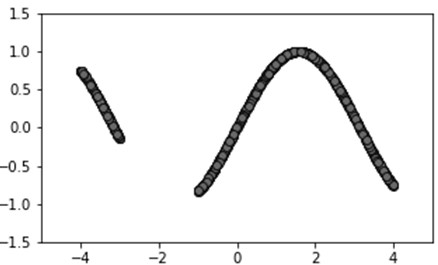}
    \caption{Epistemic uncertainty: example 1.}
    \label{fig:uc_a}
\end{subfigure}
\hfill
\begin{subfigure}{0.23\textwidth}
    \includegraphics[width=\textwidth]{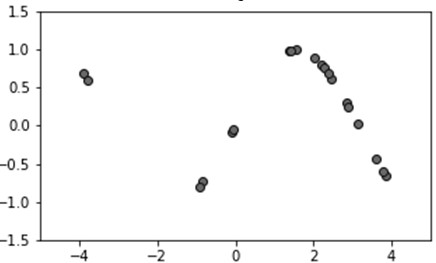}
    \caption{Epistemic uncertainty: example 2.}
    \label{fig:uc_b}
\end{subfigure}
\hfill
\begin{subfigure}{0.23\textwidth}
    \includegraphics[width=\textwidth]{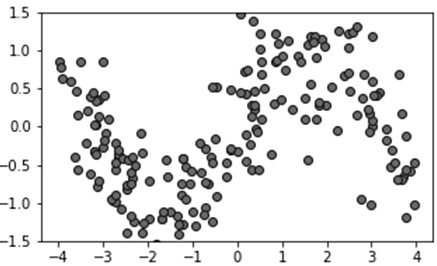}
    \caption{Aleatoric uncertainty: homoscedastic.}

    \label{fig:uc_c}
\end{subfigure}
\hfill
\begin{subfigure}{0.24\textwidth}
    \includegraphics[width=\textwidth]{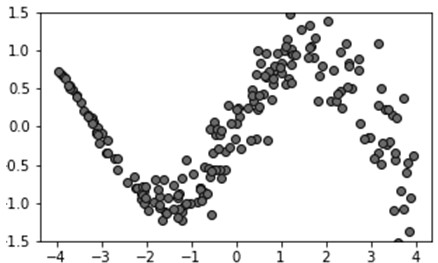}
    \caption{Aleatoric uncertainty: heteroscedastic.}
    \label{fig:uc_d}
\end{subfigure}

\caption{Examples of uncertainty types.}
\label{fig:uncertainty_types}
\end{figure}

We consider two types of uncertainty \cite{uncertainty}. The first, the \textit{epistemic} \textit{uncertainty}, expresses the uncertainty of the model caused by a lack of training data in certain parts of the domain. The epistemic uncertainty is ``reducible'' since it can be reduced by adding more samples to the training dataset. Figs.~\ref{fig:uc_a} and~\ref{fig:uc_b} show two examples. The second type of uncertainty, the \textit{aleatoric uncertainty}, is caused by observational noise and is often symbolized by $\sigma$. This type of uncertainty does not decrease by introducing more data. In practice, many models make the implicit assumption that the aleatoric noise is constant or \textit{homoscedastic} (Fig.~\ref{fig:uc_c}). In reality, the aleatoric noise likely varies across the domain, a phenomenon known as \textit{heteroscedasticity} (Fig.~\ref{fig:uc_d}).


\subsection{Estimating epistemic uncertainty with Bayesian neural networks}\label{subsec:bnn}
Given a training set of inputs $\bm{X}$ with observed targets $\bm{Y}$, the maximum likelihood estimate (MLE) of a \textit{deterministic} NN's (called $\mathcal{H}$) weights $\bm{\omega}$ can be determined by maximizing the probability $p(\bm{Y}|\bm{X},\bm{\omega}, \mathcal{H})$. The prediction $y^*=\mathcal{H}(\bm{x}^*, \bm{\omega})$ is a point estimate. Deterministic NNs are capable function approximators. Unfortunately, they suffer from overfitting when they are trained on relatively small datasets and thus require regularization measures. Deterministic NNs do not realize when a sample $\bm{x}^*$ is far from the training data $\bm{X}$ and therefore have no knowledge of the uncertainty of their corresponding prediction.

When instead determining the maximum a posteriori (MAP) distribution of the weights $\bm{\omega}$ given a training set $[\bm{X, Y}]$, we introduce stochasticity to the NNs. In the Bayesian approach, we use the Bayesian rule to reformulate the distribution $p(\bm{\omega}|\bm{X},\bm{Y},\mathcal{H})$ as:

\begin{equation}
 p(\bm{\omega}|\bm{X},\bm{Y},\mathcal{H}) = \frac{p(\bm{Y}|\bm{X}, \bm{w},\mathcal{H}).p(\bm{\omega}|\mathcal{H})}{p(\bm{Y},\bm{X})} \text{ or posterior = }\frac{\text{likelihood x prior}}{\text{evidence}}
\end{equation}

\begin{figure}
\centering
\begin{subfigure}{0.3\textwidth}
    \includegraphics[width=\textwidth]{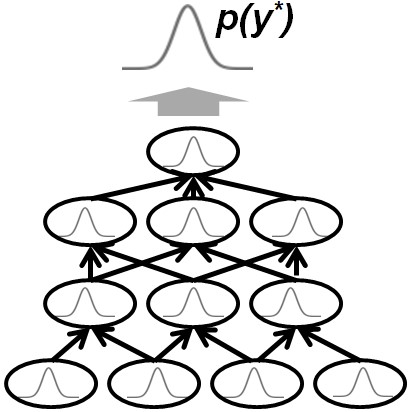}
    \caption{Bayesian neural network with distributions instead of scalars for its weights and output.}
    \label{fig:BNN_1}
\end{subfigure}
\hfill
\begin{subfigure}{0.3\textwidth}
    \includegraphics[width=\textwidth]{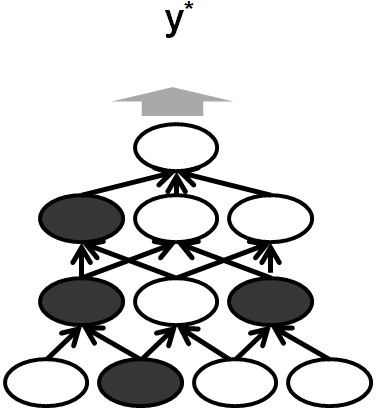}
    \caption{Dropout mechanism, with a random number (dark) of nodes ignored during training, and, for BNNs, during inference.}
    \label{fig:BNN_2}
\end{subfigure}
\hfill
\begin{subfigure}{0.3\textwidth}
    \includegraphics[width=\textwidth]{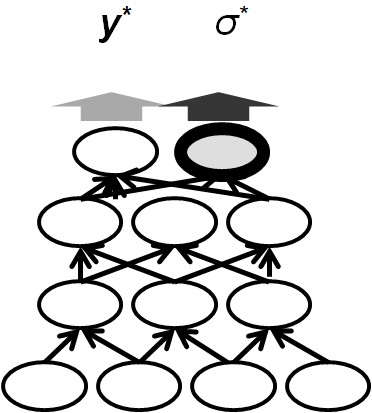}
    \caption{Additional final layer (gray with thick border) for predicting the aleatoric uncertainty.}
             \begin{minipage}{.1cm}
            \vfill
            \end{minipage}
    \label{fig:BNN_3}
\end{subfigure}

\caption{Symbolic representations of the neural networks.}
\label{fig:BNN}
\end{figure}

Interestingly, the likelihood is identical to the above MLE problem. \textit{Inference} consists of marginalizing the likelihood over $\bm{\omega}$ to obtain a target prediction $p(y^{*})$ for a given $\bm{x}^*$ (the $\mathcal{H}$ is dropped for notational simplification):
\begin{equation}\label{eq:inference}
p(y^{*}|\bm{x^{*}},\bm{X},\bm{Y}) = \int p(y^*|\bm{x}^*,\bm{\omega}).p(\bm{\omega}|\bm{X},\bm{Y}).d\bm{\omega}
\end{equation}
In contrast to the point estimate $y^{*}$ in the MLE approach, $p(y^{*}|\bm{x^{*}},\bm{X},\bm{Y})$ is a distribution, as shown in Fig.~\ref{fig:BNN_1}. The mean of this distribution, however, can be considered as a point estimate, while its variance provides a measure of the epistemic uncertainty of that estimate.

The analytical solution for the posterior distribution $p(\bm{\omega}|\bm{X},\bm{Y})$ \cite{MacKay} requires strong assumptions and has high computational complexity. Markov chain Monte Carlo sampling is not practically feasible. Therefore, we approximate $p(\bm{\omega}|\bm{X},\bm{Y})$ as a closed function $q_\theta(\bm{\omega})$ over the domain $\bm{\omega}$ that is parameterized by $\theta$. For this approximation, we minimize the Kullback-Leibler (KL) divergence between $p(\bm{\omega}|\bm{X},\bm{Y})$ and $q_\theta(\bm{\omega})$ \cite{summerschool}:
\begin{equation}
\min \, KL\left(q_\theta(\bm{\omega})||p(\bm{\omega}|\bm{X}, \bm{Y})\right)=\int q_\theta(\bm{\omega}).
\text{log}\frac{q_\theta(\bm{\omega})}{p(\bm{\omega}|\bm{X}, \bm{Y})}
\end{equation}

This minimization problem can be rewritten as maximizing the \textit{evidence lower bound} (ELBO) \cite{summerschool}:
\begin{equation}\label{eq:ELBO}
\max \,\text{ELBO}=\mathbf{E}_{q_\theta(\bm{\omega})}\text{log}\, p(\bm{X},\bm{Y}| \bm{\omega})-KL\left(q_\theta(\bm{\omega})||p(\bm{\omega})\right)
\end{equation}
The determination of the maximum of the first term is the familiar MLE approach. The second term ensures that the approximate posterior distribution $q_\theta(\bm{\omega})$ is similar to the prior distribution $p(\bm{\omega})$. There is no analytical solution for the (derivative of) the first term $\mathbf{E}_{q_\theta(\bm{\omega})}\text{log}\, p(\bm{X},\bm{Y}| \bm{\omega})$. The density function $q_\theta(\bm{\omega})$ in $\frac{\partial}{\partial\theta}\int q_\theta(\bm{\omega})\text{log}\, p(\bm{X},\bm{Y}| \bm{\omega}).d\bm{\omega}$ parameterized by $\theta$ precludes typical Monte Carlo (MC) integration. The so-called \textit{reparameterization trick} \cite{repartrick} offers an alternative solution \cite{galthesis}. By expressing $\bm{\omega}$ as a deterministic function $g(\epsilon,\theta)$, the reparameterization trick replaces sampling the dependent $\bm{\omega}$ in $q_\theta(\bm{\omega})$ by sampling the unconditional parameter $\epsilon$ in $\mathcal{N}(0, I)$. This approach is known as \textit{stochastic variational inference}. The weights $\omega$ in the NN are often assumed to have a Gaussian distribution $\mathcal{N}(\mu, \epsilon)$ so that $\omega = \mu+\sigma.\epsilon$. However, the number of learned parameters is doubled, and the code is relatively complex, which are two serious disadvantages of this approach.

\textit{Dropout} \cite{hintonDrop} is a well-known regularization technique for preventing NN overfitting (see Fig.~\ref{fig:BNN_2}). This approach involves randomly ignoring or dropping some of the NN's hidden nodes, as well as their incoming and outgoing connections, with a probability $p$ for each batch during training. This is achieved by multiplying the output of each node by a parameter $\epsilon$, which is sampled from a Bernoulli distribution with probability $p$. As a result, dropout resembles training an ensemble of NNs with lower capacities. It can be shown that maximizing the ELBO is equal to minimizing the dropout loss function of the NN with an additional L2 regularizer \cite{galDrop}. Standard NNs with dropout are easy to implement and can be used as BNNs without the need to learn new parameters, thus addressing the two aforementioned disadvantages. Often, the potential of dropout has not been fully exploited, as dropout layers are found only between convolutional layers in CNNs and only after the inputs and last LSTM layer in LSTMs. Applying dropout to the inner-product layers (kernels)~\cite{CNN} in CNNs and to all eight weight matrices in LSTM cells \cite{RNN} increases the power of the dropout. We have chosen the dropout approach for our BNNs (which include both LSTMs and CNNs; see Subsection~\ref{subsec:cnnlstm}).

Once a BNN model has been trained using a standard NN with dropout and L2 regularization, the model is ready for inference. We use MC sampling and perform $T$ stochastic forward passes on our trained model with the dropout mechanism.
The mean of the MC samples can be used to estimate the mean of Eq.~\eqref{eq:inference}:
\begin{equation}\label{eq:mean}
\mathbf{E}_{p(y^*|\bm{x}^*,\bm{X},\bm{Y})}[y^*]\approx\frac{1}{T}\sum_t\mathcal{H}(\bm{x}^*, \bm{\hat{\omega}})
\end{equation}
in which sampling $\epsilon$ from $\mathcal{N}(0, I)$ leads to indirect samples of $\bm{\hat{\omega}}$ from $q_\theta(\bm{\omega})$. Equation~\eqref{eq:var} can be used to calculate the variance:
\begin{equation}\label{eq:var}
\text{Var}_{p(y^*|x^*,\bm{X},\bm{Y})}[y^*] \approx\sigma^2+\frac{1}{T}\sum_t\mathcal{H}(\bm{x}^*, \bm{\hat{\omega}})^2-\left(\frac{1}{T}\sum_t\mathcal{H}(\bm{x}^*, \bm{\hat{\omega}}) \right)^2
\end{equation}
The last two terms $\frac{1}{T}\sum_t\mathcal{H}(\bm{x}^*, \bm{\hat{\omega}})^2-\left(\frac{1}{T}\sum_t\mathcal{H}(\bm{x}^*, \bm{\hat{\omega}}) \right)^2$ are the sample variances of the $T$ stochastic forward passes. This sample variance is regarded as the epistemic, or model, uncertainty and decreases as the size of the training dataset increases. Thus, we have shown how BNNs determine the epistemic uncertainty of each prediction.

\subsection{Estimating the heteroscedastic aleatoric uncertainty}\label{subsec:has}
The $\sigma$ in Eq.~\eqref{eq:var} is the aleatoric uncertainty. This uncertainty plays no role and hence rarely appears in the loss functions of most NNs because of the implicit homoscedasticity assumption, which results in a constant $\sigma$ across the entire domain. However, the heteroscedasticity assumption requires that an individual $\sigma_n$ be learned for each sample $\bm{x}_n$. By doubling the last dense layer in the NN, we can learn $\sigma_n$ in an unsupervised manner \cite{uncertainty}, as shown in Fig.~\ref{fig:BNN_3}. Next, we need to restore the complete MLE loss function. Therefore, we take the log of the product of the Gaussian errors, and instead of deleting the constant $\sigma$, we leave the learned $\sigma_n$ in the equation and obtain (ignoring the regulation term):
\begin{equation}\label{eq:hetero}
L= \min \frac{1}{N} \sum \frac{1}{2\sigma_n^2}(\bm{y}_n-\mathcal{H}(\bm{x}_n))^2 +\frac{1}{2}\text{log}\,\sigma_n^2
\end{equation}
For samples in the noisier parts of the domain, the model predicts higher aleatoric uncertainties $\sigma_n$, reducing their effect on the total loss $L$. This process, known as \textit{loss attenuation}, should result in more accurate predictions. The second term in Eq.~\eqref{eq:hetero} is a regularizer, which prevents the model from always predicting a high value of $\sigma_n$ to minimize $L$. It should be noted that with the homoscedasticity assumption, $\sigma_n$ is constant, and Eq.~\eqref{eq:hetero} becomes the commonly used minimization of the sum of squared errors.

\subsection{Classification}
Subsections~\ref{subsec:bnn} and~\ref{subsec:has} focused on regression problems. Classification problems are highly similar. However, since classification NNs output probability vectors rather than scalars, their outputs require a different interpretation. A classification NN consists of a regression NN $f^{\bm{\omega}}$ that outputs a vector of \textit{logits} that is normalized through an additional layer with a \textit{softmax} function, yielding the following probability vector:

\begin{equation}
\bm{p}(\bm{x}^{*})=\text{softmax}([f_1^{\bm{\omega}}, f_2^{\bm{\omega}}, ..., f_C^{\bm{\omega}}])
\end{equation}
where $C$ is the number of possible classes. The probability $p_c$ is interpreted as the probability of class $c$ relative to the other classes and is not an expression of the overall certainty of the model.\\
Similar to regression, our BNN learns a distribution of the weights $p(\bm{\omega}|\bm{X},\bm{Y},\mathcal{H})$ by maximizing the ELBO and approximating with dropout regularization. Information theory concepts are then used to measure the epistemic and aleatoric uncertainties \cite{galthesis}. We interpret the entropy (the average amount of information contained in a distribution) of the predicted probability vector as a measure of the prediction's total uncertainty. The entropy of $\bm{p}(\bm{x}^{*})$ is given by:
\begin{equation}\label{eq:entropy}
\mathbb{H}_{\bm{p}(y^*|\bm{x}^{*})} = - \sum_{y^*=c}p(y^*=c|\bm{x}^*).\text{log}\,p(y^*=c|\bm{x}^*)
\end{equation}
where $c$ presents all possible classes. We use the MC approximation to reduce bias by sampling $T$ times:
\begin{equation}
(y^*=c|\bm{x}^*) \approx \frac{1}{T}\sum_t p^{\hat{\bm{\omega}}_t}(\bm{x}^*)_c
\end{equation}
where the weights $\hat{\bm{\omega}}_t$ are sampled from $q_\theta(\bm{\omega})$ and $p^{\hat{\bm{\omega}}_t}(\bm{x}^*) = \text{softmax}\left(f^{\hat{\bm{\omega}}_t}(\bm{x}^*) \right)$.
$\mathbb{H}$ is large when the resulting predictive vector is nearly uniform, which reflects a high uncertainty, as shown in the first three columns of Table~\ref{table:evaluationmethods}.
The predictive entropy does not enable us to differentiate between the aleatoric and epistemic uncertainties. The mutual information $\mathbb{I}$ adds an expectation term to the predictive entropy $\mathbb{H}$:
\begin{equation}
\mathbb{I}(y^*,\bm{\omega}|\bm{x}^*)=\mathbb{H}_{\bm{p}(y^*|\bm{x}^{*})}-\mathbb{E}_{p(\bm{\omega})} [\mathbb{H}_{p(y^*|\bm{x}^*,\bm{\omega})}]
\end{equation}
The second term $\mathbb{E}_{p(\bm{\omega})} [\mathbb{H}_{p(y^*|\bm{x}^*,\bm{\omega})}]$ can be approximated by MC sampling:
\begin{equation}
\mathbb{E}_{p(\bm{\omega})} [\mathbb{H}_{p(y^*|\bm{x}^*,\bm{\omega})}]\approx \frac{1}{T}\sum_t p^{\hat{\bm{\omega}}_t}(\bm{x}^*)_c
\text{log}\,p^{\hat{\bm{\omega}}_t}(\bm{x}^*)_c
\end{equation}
The weights $\hat{\bm{\omega}}_t$\ are sampled from $q_\theta(\bm{\omega})$, which is used to approximate the posterior. When the second term $\mathbb{E}_{p(\bm{\omega})} [\mathbb{H}_{p(y^*|\bm{x}^*,\bm{\omega})}]$ is equal to the first term $\mathbb{H}$ and $\mathbb{I}$ is zero, all functions have the same $y^*$ for a given input $\bm{x}^*$. The model has a large certainty, and the mutual information can be used to measure the epistemic (model) uncertainty. As the predictive entropy represents the total uncertainty, the difference between the predictive entropy and the mutual information $\mathbb{H} - \mathbb{I}=\mathbb{E}_{p(\bm{\omega})} [\mathbb{H}_{p(y^*|\bm{x}^*,\bm{\omega})}]$ represents the aleatoric uncertainty, as seen from the examples in Table~\ref{table:evaluationmethods}.

\begin{table}[!ht]
\begin{center}
\caption{Interpretation of the predictive entropy ($\mathbb{H}$) and mutual information in terms of the total, epistemic and aleatoric uncertainties  ($\mathbb{I}$)}\label{table:evaluationmethods}
\scalebox{0.93}{
\begin{tabular}{lllll}
  \toprule
   MC function draws & $y^*$ &$\mathbb{H}$: total & $\mathbb{I}$: model & $\mathbb{H}$-$\mathbb{I}$: observ-\\
   $p^{\hat{\bm{\omega}}_t}(\bm{x}^*)$&&  Predictive &  Uncertainty&
ational noise \\
    &&Uncertainty&(Epistemic)&(Aleatoric)\\
  \midrule
  \{(1, 0), (1, 0), ... (1, 0), (1, 0)\}  & (1, 0)  & 0 (low) & 0 (low) &  0 (low) \\
  \{(0.5, 0.5), ..., (0.5, 0.5)\} &(0.5, 0.5) & 0.69 (high) & 0 (low) & 0.69 (high)\\

    \{(1, 0), (0, 1), ..., (1, 0), (0, 1)\} &(0.5, 0.5) & 0.69 (high) &0.69 (high)&  0 (low)\\

  \bottomrule
\end{tabular}
}
\end{center}
\end{table}

The aleatoric uncertainty is determined after the model is trained and thus cannot be used for loss attenuation, as in Subsection~\ref{subsec:has}. Moreover, the loss curve $L$ does not depend on $\sigma$. When the MLE procedure is used for classification tasks, the equivalent of Eq.~\eqref{eq:hetero} becomes the \textit{cross-entropy loss} (CEL):
\begin{equation}
L=\min \sum_n^N\left(-f^{\bm{\omega}}_c(\bm{x}_n) + \text{log} \sum_{c'}^C exp(f^{\bm{\omega}}_{c'}(\bm{x}_n)) \right) = \min\sum_n^N\text{CEL}
\end{equation}
where $c$ is the observed class of $\bm{x}_n$ and $c'$ are all the possible classes. To enable loss attenuation, we place a Gaussian distribution with variance $\sigma_n^2$ over the results of the regression part of the classification model before passing them to the softmax layer \cite{uncertainty}:

\begin{equation}
\hat{\bm{a}}_n\sim\mathcal{N}\left(f^{\bm{\omega}}(\bm{x}_n),\sigma_n^2\right)
\end{equation}
\begin{equation}
\hat{p}^{\bm{\omega}}(\bm{x}_n) = \text{softmax}(\hat{\bm{a}}_n)
\end{equation}
The MLE loss function has no analytical solution, and we thus again use MC sampling:
\begin{equation}\label{eq:class_lossatt}
\hat{\bm{a}}_{n, t}=f^{\bm{\omega}}(\bm{x}_n)+\sigma_n\epsilon_t,\;\;\epsilon_t\sim\mathcal{N}(0, I)
\end{equation}
This sampling is computationally inexpensive, as the regression model inference $f^{\bm{\omega}}(\bm{x}_n)$ does not need to be repeated. Finally, we determine an attenuated loss function that can be written in terms of the CEL:
\begin{equation}
L = - \sum_n^N\text{log}\frac{1}{T}\sum_t^T \text{exp}\left(-CEL_{n, t}\right)
\end{equation}
In our models, we used a more numerically stable variant of this loss function~\cite{kyle}. With our terminology and above notations, we can formulate this loss function as:
\begin{equation}
L=\min \sum_n^N \left(CEL_n \left(1 + \frac{1}{T}\sum_t^T \text{ELU}(CEL_{n, t} - CEL_n\right) + \text{exp}(\sigma_n^2)-I_N\right)
\end{equation}
where ELU is the \textit{Exponential Linear Unit} activation function, which is given by:
\begin{equation}
\text{ELU}(x) = \begin{cases}
 x \text{ for } x > 0, \\
\alpha . \text{exp}(x) - 1 \text{ for } x \leq 0
 \end{cases}
\end{equation}

\subsection{LSTM vs. CNN}\label{subsec:cnnlstm}
Among NNs, researchers often prefer LSTMs for predictive process monitoring problems. However, an increasing number of papers (e.g., \cite{bai}) have identified CNNs as a quick alternative for time series and sequence problems, a hypothesis confirmed by \cite{cnnlstm} for process outcome prediction. As the previously described uncertainty techniques can be implemented as extensions to plain-vanilla CNN and LSTM architectures, we integrated these techniques into CNNs and LSTMs to better understand the relative merits of both models in our experiments.

\subsection{Research questions}\label{subsec:goals}
After establishing an understanding of the theory, we can now divide our initial goal of evaluating uncertainty quantification techniques in predictive process monitoring into four research questions.
\begin{enumerate}[{RQ}1:]

\item What is the quality of our uncertainty estimates in predictive process monitoring tasks?
   \begin{enumerate}[a.]
        \item \textbf{Correlation with accuracy}: Do the calculated uncertainties indicate the accuracy of the point estimates (remaining time), and can these uncertainties be used to discriminate between good and bad estimates?
        \item \textbf{Classification}: Are the above uncertainties also valid for classification problems (outcome prediction)?
        \item \textbf{Confidence intervals}: How can we use these uncertainties to build confidence intervals for the NNs' predictions?
        \item \textbf{Early predictions}: How does the uncertainty behave during the early phases of (incomplete) cases?
    \end{enumerate}

\item How do the following techniques affect the overall quality of the point estimates:
    \begin{enumerate}[a.]
        \item \textbf{Heteroscedasticity}: Learning the observation noise for individual samples ($\sigma_n$) enables loss attenuation. Does this knowledge also lead to more accurate point predictions?
        \item \textbf{Dropout}: We added dropout to heteroscedastic NNs. How well does dropout perform in isolated, non-Bayesian contexts?
        \item \textbf{BNN}: We added L2 regularization and applied MC sampling ($T$ stochastic forward passes) to our heteroscedastic NNs with dropout and determined the point estimates by calculating the mean (Eq.~\eqref{eq:mean}). Are these predictions more accurate than those of deterministic models?
        \item \textbf{CNN/LSTM/base case}: How do CNNs compare to LSTMs, and how do both compare to baseline models?
    \end{enumerate}
\item How are the training and inference computation times related, and are uncertainty-enhanced NNs slower than otherwise equivalent deterministic models?

\item What is the applicative value of the knowledge of the uncertainties of the predictions in the field of the predictive process monitoring?
\end{enumerate}

\section{Experimental setup} \label{sec: eval}

Our experimental setup was based on answering the above research questions rather than obtaining the highest possible accuracy or optimizing the underlying NNs.

\subsection{Datasets}

We used five publicly available real-world datasets. BPIC\_2012\footnote{\url{https://data.4tu.nl/articles/dataset/BPI\_Challenge\_2012/12689204}} includes loan application processes at a Dutch bank.  BPIC\_2015\footnote{\url{https://data.4tu.nl/collections/BPI\_Challenge\_2015/5065424}} is a collection of building permit applications in five Dutch municipalities that we considered as five separate datasets. BPIC\_2017\footnote{\url{https://data.4tu.nl/articles/dataset/BPI\_Challenge\_2017/12696884}} is a richer, cleaner and larger set of logs of Dutch bank loan application processes. BPIC\_2019\footnote{\url{https://data.4tu.nl/articles/dataset/BPI\_Challenge\_2019/12715853}} has a similar size to the above dataset; however, its cases are much shorter and related to purchase order handling processes. BPIC\_2020\footnote{\url{https://data.4tu.nl/collections/BPI\_Challenge\_2020/5065541}} is a set of five smaller datasets sourced from a travel administration at a university, three of which are suitable for our experiments (see~\cite{bias} for the applicability of the datasets). These datasets include records of processes covering prepaid travel costs and requests for payment (``Payment''), travel permits (``Permit'') and expense claims (``Travel Costs'').

\subsection{Preprocessing and targets}\label{subsec:preprocessing}
We followed the preprocessing steps described in \cite{bias}. Clear chronological outliers were deleted.
The test sets included the 20\% most recent cases, determined by starting time.
To prevent data leakage (where the duration and outcome of cases are only known at the end) and mutual influence, we removed all cases in the training sets that were still running at the same time as the earliest cases in the test sets. Bias in terms of the number of open cases and the average duration of cases at the beginning and end of the test sets was removed. To remove this bias, we removed some of the cases with the longest durations. For reproducibility, the code for these steps is publicly available\footnote{\url{https://github.com/hansweytjens/predictive-process-monitoring-benchmarks}\label{footnote:github}}. To assess the effect of the training set size on the results, we used increasing shares (between 1\% and 100\%) of randomly selected cases from the total training set for the three largest datasets (BPIC\_2012, BPIC\_2017, and BPIC\_2019). The results of the remaining datasets (BPIC\_2015 and BPIC\_2020) were ranked according to training set size to allow for similar insights. Table~\ref{tab:data} provides an overview of the main characteristics of the datasets.\\

\begin{table}
\centering
\caption{Statistics of the used datasets after preprocessing.}\label{tab:data}
\scalebox{0.9}{
\begin{tabular}{lcccccccc}
\toprule
Dataset& Avg.  & \% &Avg.  & Share of     &Training         &Test    &Levels\\
                &Nr. of & Approved$^*$ & Nr. of       & Events                          & Events             & Events & \\
                                & Days$^*$ & &Events$^*$     & Used               &            &              &  \\
\midrule
BPIC\_2012  &6.4&18.2& 18.4 & 1\%&1,198 & 34,549  & 26 \\
                            &| &| & |                                            &  2\%&1,954 & |  & | \\
                            &...&...& ...                                            & ...     & ...&...&...\\
                            &| &|& |                                           &  50\% &58,818 &| &|\\
                            &$\downarrow$ &$\downarrow$&  $\downarrow$                                          &  100\%& 117.546& $\downarrow$& $\downarrow$\\
\midrule
BPIC\_2017 & 19.8&74.9&39.3  &1\% & 7,639 & 228,421 &  27\\
                        &| & |   &    |                                         & 2\% &16,705  & | & | \\
                           &...&...&   ...                                          & ...     & ...&... &...\\
                            &|&|&  |                                           & 50\% &403,034&| &|\\
                            &$\downarrow$  &$\downarrow$ &       $\downarrow$                                   & 100\% &805,809 &$\downarrow$&$\downarrow$\\
\midrule
BPIC\_2019 &52.14&N/A&                                  6.3                & 1\% &5,323&227,886&42\\
                       &| &|&          |                                       & 2\% &11,180 & | & |\\
                       &...&...&  ...                                              & ...     &... &...&...\\
                      & | &|&  |                                              & 50\% & 268,471 &|&|\\
                      & $\downarrow$ &$\downarrow$& $\downarrow$                                               & 100\% &  538,293&$\downarrow$&$\downarrow$\\
\midrule
BPIC\_2015:&&&                                                                    & & & & \\
Municip. 2&112.6&N/A &    53.1                                                                  &100\% & 20,581 & 7,486 &237 \\
Municip. 4&105.8&|&                    47.0                            &| &25,101 & 8,839 & 249\\
Municip. 1&70.3&|&   43.2 &|  &28,154& 9,819& 247 \\
Municip. 5&85.7&|&     52.6                                              & |& 34,308& 10,262&257\\
Municip. 3&48.3&$\downarrow$&       43.4                                 &$\downarrow$ &34,967& 11,191& 255 \\
\midrule
BPIC\_2020:&&&                                                                    & & &  &  \\
Travel Costs&27.5&N/A& 9.2                                                 & 100\%& 10,809& 3,714&   31\\
Payments&8.8&|& 5.3                                                     & |&23,295 &6,200 &18\\
Permits&64.9&$\downarrow$&13.9      & $\downarrow$& 51,174& 22,145& 52 \\

\bottomrule
\multicolumn{8}{l}{\small * on a case basis, for training and test set.} \\
\end{tabular}
}
\end{table}
Additional preprocessing included the addition of one synthetic feature, the ``elapsed time'' (since the start of the case), which was calculated using the event time stamps. The features ``activity'' and ``elapsed time'' were the only features used for our experiments. The range feature ``elapsed time'' was standardized. We did not cap the number of levels of the categorical variable ``activity'', as nonfrequent labels could lead to uncertain estimates. To prepare the labels as inputs for the embedding layer in the NNs, we mapped the labels to integers. We extracted all feasible prefixes from the dataset and standardized these prefixes to a \textit{sequence length} of ten by padding the shorter sequences and truncating the longer sequences. Our main target, the remaining execution time, was represented as the fractional number of days remaining until the end of the case. For the classification task, the target was defined as ``approved'' in both the BPIC\_2012 and BPIC\_2017 datasets (for details, see~\cref{footnote:github}).

\subsection{Models, metrics, training and environment}\label{subsec:models}

Our NN model consisted of two convolutional or LSTM layers followed by three dense layers (plus a softmax function in the classification case). Dropout was implemented as described in Subsection~\ref{subsec:bnn}. The mean absolute error (MAE) was the metric for the remaining time task. We used the transition system-based method~\cite{aalst} as a baseline in our remaining time (regression) experiments to evaluate the absolute performance of our models. Because the BPIC\_2012 and BPIC\_2017 datasets are both ill-balanced with respect to our chosen target (see Table~\ref{tab:data}), we used the area under the curve receiver operating characteristic (AUC\_ROC) as a metric for measuring the quality of the outcome predictions (classification).
All experiments were carried out in Python/PyTorch, and the code can be found on GitHub\footnote{\url{https://github.com/hansweytjens/uncertainty}}.
\subsection{Estimating the epistemic, aleatoric and total uncertainties}

\begin{table}
\centering
\caption{Experimental settings.}\label{tab:experiments}
\begin{tabular}{ll}
\toprule
Setting & Value \\
\midrule
Nr. runs per experiment (for larger datasets)&  10 \\
\hspace{98pt}(for smaller or small fractions of datasets) & 50 \\
\midrule
Nr. MC samples BNN ($\mathcal{H}(\bm{x}^*, \bm{\hat{\omega}})$ in Equations~\ref{eq:mean} and ~\ref{eq:var}) & 50 \\
Nr. MC samples probability vector ($p^{\hat{\bm{\omega}}_t}(\bm{x}^*)$ in Equation~\ref{eq:entropy}) & 50 \\
Nr. MC samples softmax ($\hat{\bm{a}}_{n, t}$ in Equation~\ref{eq:class_lossatt}) & 20 \\
\midrule
Max. sequence length (padding for shorter prefixes)&10\\
Optimizer&Adam\\
Max. nr. epochs (large - small datasets) &20-400\\
Early stopping &yes\\
Dropout parameter $p$&0.1\\
\bottomrule

\end{tabular}
\end{table}

For each prefix in the test set, we performed MC sampling by conducting $T=50$ stochastic forward passes; each pass used a different $\epsilon$ for each weight $\omega$ in the model. According to Eq.~\eqref{eq:mean}, the mean of the results could be calculated to determine the point estimate of the BNN. According to Eq.~\eqref{eq:var}, the variance of the results indicated the epistemic uncertainty.
The additional final dense layer in the models was used to determine the per-point aleatoric uncertainty, which became part of the adapted loss functions, as in Eq.~\eqref{eq:hetero}. The total uncertainty in Subsection~\ref{subsection:different} and Section~\ref{sec:working} was determined as the summation of the epistemic and aleatoric uncertainties, as suggested by Eq.~\eqref{eq:var}. For the classification task, we used 50 samples to calculate the total uncertainty according to the entropy in Eq.~\eqref{eq:entropy} and pushed 20 samples from $\hat{\bm{a}}_{n, t}$ (as per Eq.~\eqref{eq:class_lossatt}) through the softmax function during each forward pass of the model. Each experiment was repeated 10-50 times, inversely correlated to the size of the dataset. The published metrics are the averages of these experiments. These metrics and other parameter settings are summarized in Table~\ref{tab:experiments}.

\section{Results}\label{sec:results}

\subsection{Uncertainty estimates (RQ1)}\label{subsection:different}
We investigated to what extent the total uncertainty estimates corresponded with the quality of the predictions and the reliability of the confidence intervals. We used Bayesian CNNs with a p=5\% dropout probability and built-in aleatoric uncertainty learning. We refer to these models as BNNs in this section for simplicity.
\paragraph{Certainty of the predictions correlates strongly with accuracy (RQ1.a)}

In Fig.~\ref{fig:uncertainty}, each column represents a dataset\footnote{The results in this and the following figures differ from those in \cite{uncertainty_WW}, as we used unbiased datasets without data leakage~\cite{bias}, did not include concrete dropout and did not apply early stopping based on a validation set. Instead, we ran our models until the learning curve flattened.}. The horizontal axes in the graphs of the BPIC\_2012, BPIC\_2017 and BPIC\_2019 datasets show the proportion of samples in the training set that were effectively used for training, with the values increasing from left to right. The two right-hand columns include the five or three subdatasets of the BPIC\_2015 and BPIC\_2020 datasets ranked by size. The MAEs of the models are displayed on the vertical axes. The MAEs of the BPIC\_2015 and BPIC\_2020 datasets were normalized, with their base cases set to one. We sorted the predictions of the test set based on the estimated uncertainties. We then applied different retention thresholds to retain the top 100\%, 75\%, 50\%, 25\%, 10\% and 5\% of the predictions. The uncertainty and prediction quality are strongly correlated, as shown in Fig.~\ref{fig:uncertainty}. When the threshold was sufficiently high, it was possible to determine perfect predictions, as demonstrated by the 5\% lines on the BPIC\_2012 and BPIC\_2017 plots. Unfortunately, Fig.~\ref {fig:uncertainty} also shows that the quality of the uncertainty estimates decreases as the size of the dataset decreases, similar to the predictions themselves. Meaningful separation is impossible for the smallest BPIC\_2012 training sets.

\begin{figure}[!ht]
\center
\includegraphics[width=1\textwidth]{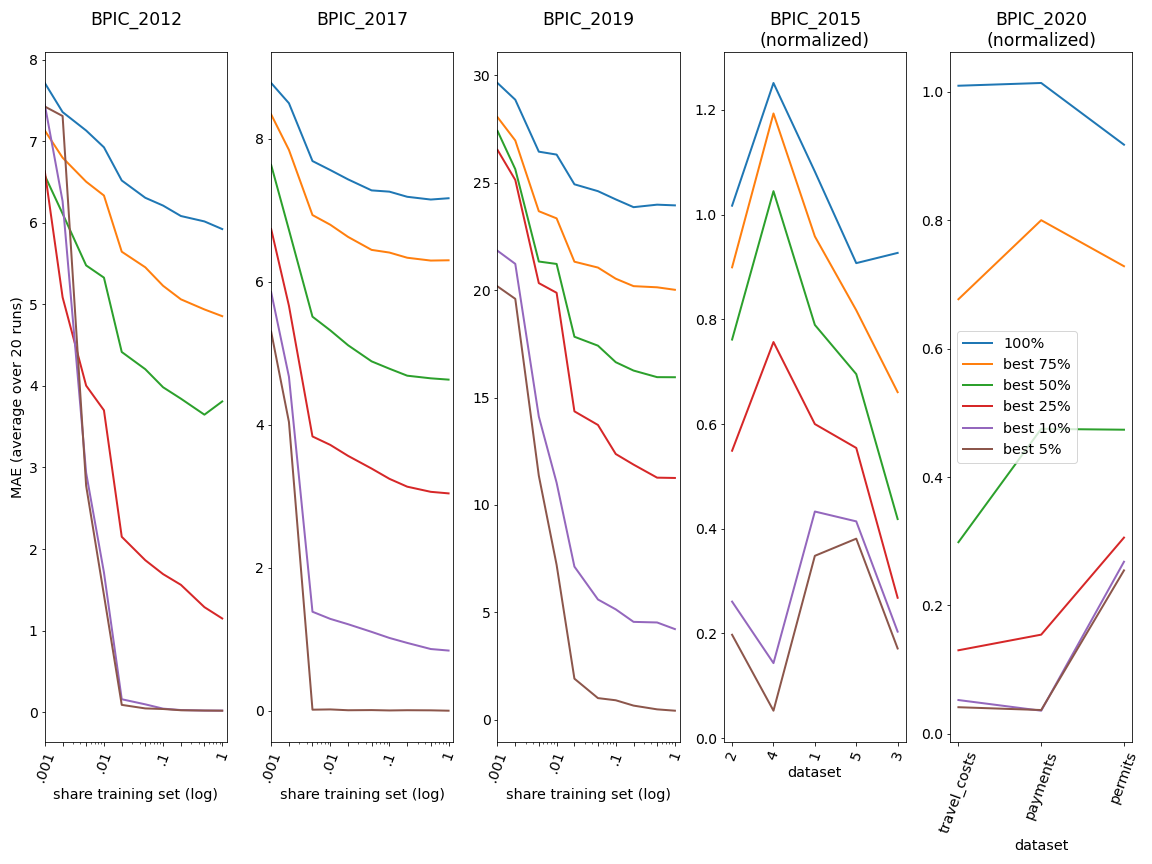}
\caption{Prediction quality (MAE) for different uncertainty thresholds. The blue line represents the complete test set, and the brown lines represent test set samples with the 5\% lowest uncertainty.} \label{fig:uncertainty}
\end{figure}

\begin{figure}[!ht]
\center
\hspace*{-1.5cm}\includegraphics[width=.8\textwidth]{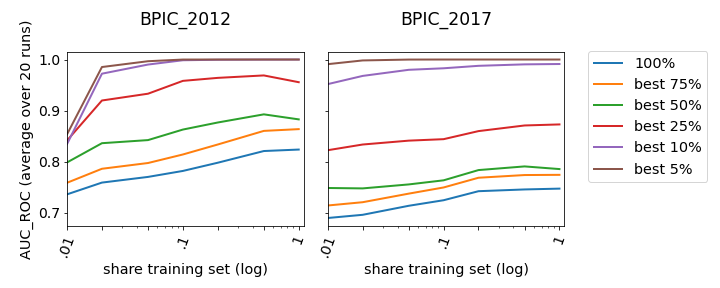}
\caption{Classification: same interpretation as Fig.~\ref{fig:uncertainty}. The uncertainty estimates correspond to the quality of the predictions.} \label{fig:classification}
\end{figure}

\paragraph{Classification (RQ1.b)}
To demonstrate how the learning uncertainty can be effectively used to determine the quality of a model's predictions in classification tasks as well, we constructed the classification equivalent of Fig.~\ref{fig:uncertainty} for the BPIC\_2012 and BPIC\_2017 datasets. The task included forecasting the ``approved'' target. The lowest blue lines in Fig.~\ref{fig:classification} represent the average AUC\_ROC over all samples in the test set for different shares of the training set. It can be clearly seen that progressively removing the worst predictions based on their uncertainty increases the average AUC\_ROC until all remaining predictions become perfect (AUC\_ROC = 1).

\begin{figure}[!ht]
\hspace*{-1.5cm}
\includegraphics[width=1.2\textwidth]{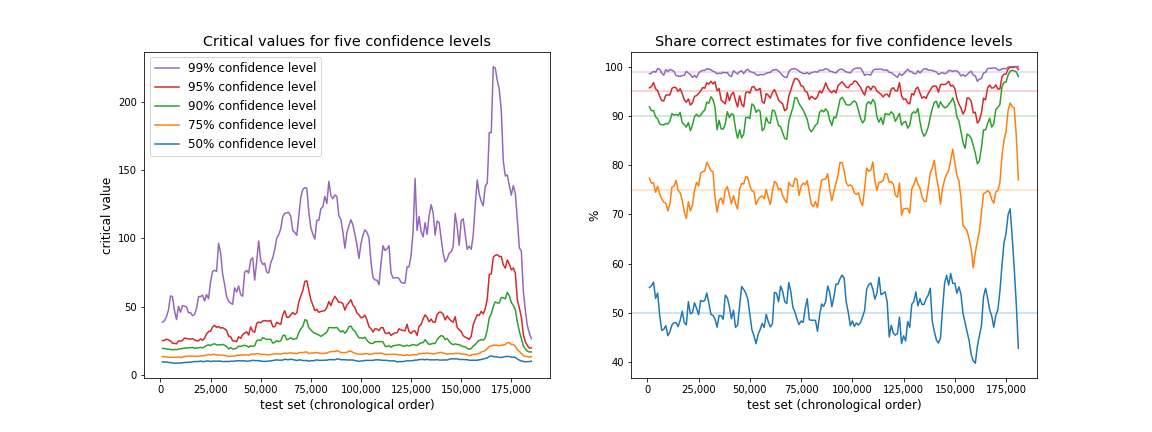}
\caption{Left: critical values for confidence levels of 50\%, 75\%, 90\%, 95\% and 99\% computed on 5,000 preceding samples every 1,000th sample in the test set. Right: Corresponding share of predicted values in the following 5,000 samples that fall within the confidence interval. The BPIC\_2017 (complete) dataset was used.} \label{fig:calibration}
\end{figure}

\paragraph{Predictions with confidence intervals (RQ1.c)}
Most machine learning techniques (with the notable exception of linear regression) do not allow confidence intervals to be built around the predictions of the model without considerable assumptions about the (unknown) distribution of these predictions. The acquired knowledge of a prediction's uncertainty allows us to build confidence intervals, as demonstrated using the BPIC\_2017 dataset. We determined the upper and lower bounds of the confidence interval by multiplying the \textit{critical value} ($z^*$ in statistics) by the uncertainty and adding/subtracting the product to/from the point estimate. We used the last 5,000 samples in the training set to compute the first critical values for confidence levels of 50\%, 75\%, 90\%, 95\% and 99\%. The critical values were calculated online using a sliding window of 5,000 samples to offset concept drift, as shown in the first graph in Fig.~\ref{fig:calibration}. The second graph in Fig.~\ref{fig:calibration} plots the proportion of predicted values that fall within the respective confidence intervals. These values oscillate around their ideal values (horizontal lines), and the confidence intervals function as expected.

\begin{figure}[!ht]
\includegraphics[width=1\textwidth]{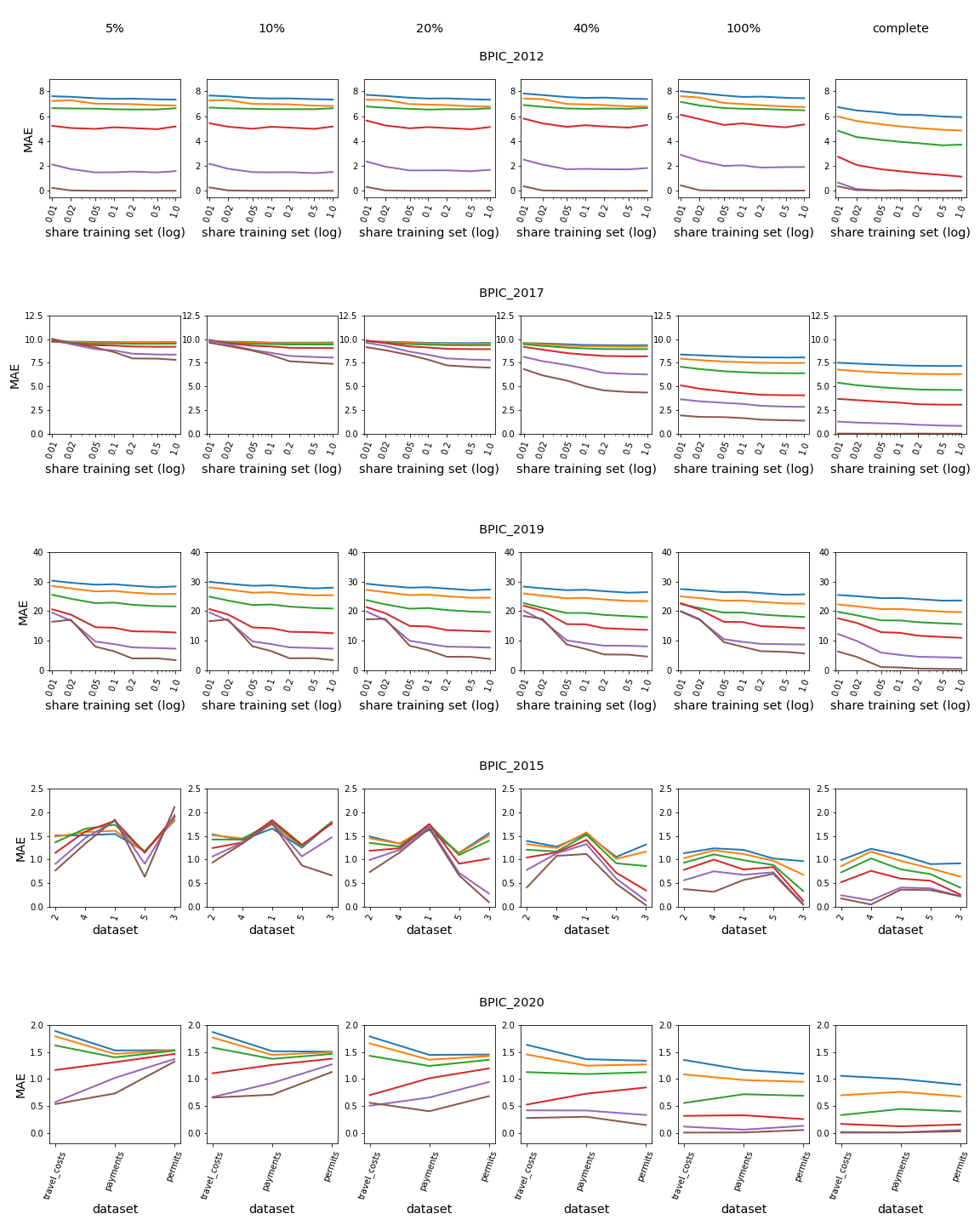}
\caption{Early predictions. The rows represent different datasets. The first column shows the predictions for prefixes with a duration of up to 5\% of the median duration of the cases in the test sets. The remaining columns contain increasingly longer prefixes. The graphs in the last column include the complete test sets and are therefore identical to those in Fig.~\ref{fig:uncertainty}. The axes and colored lines are the same as in Fig.~\ref{fig:uncertainty}.} \label{fig:earlycomplete}
\end{figure}

\paragraph{Early predictions (RQ1.d)}
The quality of the uncertainty predictions is remarkably high for early predictions. The point estimates are more accurate for longer prefixes than for shorter prefixes. This result can be observed by moving from the first to the last column in Fig.~\ref{fig:earlycomplete}. However, especially for the BPIC\_2012 and BPIC\_2019 datasets, the uncertainties during the very early stages (5-10\% of the median duration) correlate strongly with the accuracy of these point estimates. For the BPIC\_2017 and BPIC\_2015 datasets, the very short prefixes do not include enough information to correctly assess the uncertainty of the remaining time predictions. Unsurprisingly, this effect is exacerbated when the training sets are small (toward the left of the individual graphs). To calculate the results presented in this subsection, we trained our models using the complete training set, including all available prefixes, regardless of the prefix durations of the subsequent predictions. We also investigated prefix-duration bucketing, in which the models were trained on prefixes that were less than or equal to the maximum duration of the prefixes in the relevant test sets (columns). The results are presented at the same scales as in Fig.~\ref{fig:earlybuckets} in~\ref{app:bucket} and are generally inferior to the results obtained when all available prefixes in the training set were used. This result suggests that cases contain both critical events that strongly influence the remaining time until completion and less influential events that do not have such an impact. The critical events do not always occur at the beginning of a case.

\subsection{Performance of the point estimates (RQ2)}\label{subsection:overall}
\begin{figure}[!ht]
\center
\includegraphics[width=1\textwidth]{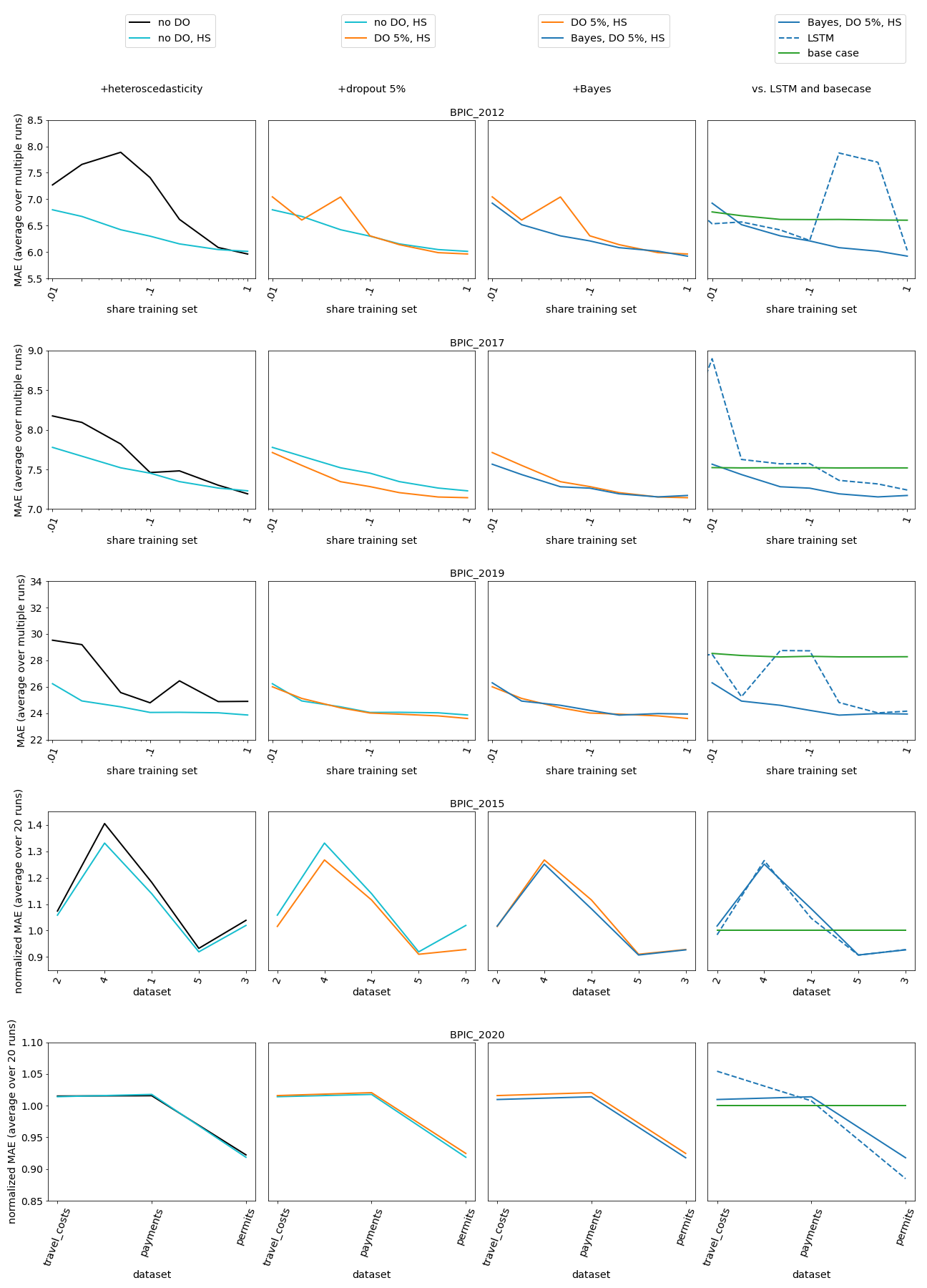}
\caption{Overall results on the complete test sets. ``no DO'': no dropout = plain-vanilla NN, ``HS'': heteroscedastic, ``DO 5\%'': 5\% dropout probability, ``Bayes'': BNN. The rows represent five different datasets, and different techniques are presented in each column. The BPIC\_2015 and BPIC\_2020 dataset results were normalized with a base case = 1.} \label{fig:summary_results}
\end{figure}

The techniques described in Subsection~\ref{subsec:goals} are used to obtain the uncertainty estimates. Theoretically, these techniques can also be used to obtain more accurate point estimates. Fig.~\ref{fig:summary_results} summarizes the results of our investigations. The rows represent different datasets, while the columns compare the techniques in a step-by-step manner, as discussed in the following four paragraphs. The scale of the MAE of the model on the vertical axis is held constant across the columns in each row.

\paragraph{Loss attenuation is very effective (RQ2.a: Fig.~\ref{fig:summary_results}: column 1)}
The predictions were significantly improved by loss attenuation (Eq.~\eqref{eq:hetero}) in the loss functions. Plain-vanilla NNs are represented by the black lines in Fig.~\ref{fig:summary_results}. The cyan lines show models that are otherwise identical, including loss attenuation. The learned per-point uncertainties appear to have successfully identified aleatoric uncertainty for certain samples before reducing the importance of these samples in the loss function to prevent the model from progressing in the wrong direction. These results differ from those in our previous paper~\cite{uncertainty_WW}, in which the use of early stopping based on a validation set effectively reduced overfitting, with little room for loss attenuation to make further improvements. However, early stopping also required the use of a validation set, thereby reducing the number of samples available for training.
Loss attenuation does not require the use of a validation set.
Thus, based on the ease of implementation and the usefulness of the learned aleatoric uncertainty (Subsection~\ref{subsection:different}), loss attenuation can be considered an important and advantageous improvement in machine learning.

\paragraph{Dropout further combats overfitting (RQ2.b: Fig.~\ref{fig:summary_results}: column 2)}
In our simulations, we used a dropout probability of 5\%. The heteroscedastic models are depicted by the cyan lines in the second column of Fig.~\ref{fig:summary_results}. Despite the effectiveness of loss attenuation, the dropout mechanism (orange lines) further reduced overfitting on some datasets, most notably BPIC\_2017 and BPIC\_2015. Thus, dropout is an important building block of BNNs.

\paragraph{Bayesian learning improves the results, even for very small datasets (RQ2.c: Fig.~\ref{fig:summary_results}: column 3)}
In the previous two steps, we used deterministic NNs to develop models with dropout (orange lines). By adding L2 as an additional regulator to the loss function (the third regulator after loss attenuation and dropout) and sampling with the dropout mechanism remaining during inference, we converted our models to stochastic BNNs, which are represented by the blue lines in column 3. These models predict distributions, and the arithmetic averages of these distributions yield point estimates. BNNs are essentially ensembles (models sampled with different weights during each MC sampling during the inference stage), which are said to outperform their constituent models in terms of accuracy. Our experiments confirmed this hypothesis for smaller datasets but found the effect to be negligible for larger datasets. BNNs also estimate the epistemic uncertainty (Section~\ref{sec:theory}), an important component of the total uncertainty that allowed us to assess the quality of the predictions.
The combination of these three regularization techniques makes the BNNs remarkably robust to overfitting. We therefore suggest constructing large-capacity BNNs and training these models over a large number of epochs. This procedure ensures that the model does not fall into the recently discovered deep double descent~\cite{doubledip} trap, in which a model's performance first decreases with increasing size and increasing number of training epochs due to overfitting before beginning to improve again.

\paragraph{CNNs outperform LSTMs, and BNNs outperform the base cases (RQ2.d: Fig.~\ref{fig:summary_results}: column 4)}
The models in the first three columns were all based on CNNs. In the fourth column, we compared the CNN-based BNNs (solid blue lines) to LSTM models with the same architecture (dotted blue lines). In most cases, the CNNs produced superior results. These results depended on the chosen architecture (number of layers, nodes, etc.) and were consistent with the findings in \cite{bai, cnnlstm}. When only a small amount of training data was available, the BNNs outperformed the base cases (green lines). For very small datasets, simpler alternative models should always be considered.

\subsection{Computation time (RQ3)}\label{subsec:speed}

\paragraph{BNNs can be trained and make predictions quickly}
BNNs are very robust to overfitting and thus facilitate the use of large architectures that have been trained for longer times than would be recommended for plain-vanilla NNs that have to rely on smaller models and measures such as early stopping to prevent overfitting. To obtain some intuition for the computation time of BNNs, we compared BNNs to their plain-vanilla equivalents of the same size and trained the models for the same number of epochs.
We found that BNNs need approximately 40\% longer training time than their plain-vanilla counterparts.

As inference requires MC sampling, BNN predictions take longer as a function of $T$, the number of samples. However, with the exception of some environments that require real-time predictions, the inference time can be ignored in most applications.

Since we can safely train large-capacity BNNs for long periods of time, the hyperparameter space to be searched decreases significantly (number of layers, number of nodes, number of epochs, etc.). Hence, in consideration of the whole training process, it might be noticeably faster to work with the slightly slower BNNs than other models.

\paragraph{CNNs are faster than LSTMs}
Unlike LSTM calculations, which are sequential, CNN calculations can be parallelized and therefore require an order of magnitude less time to complete. The implementation of dropout in LSTM cells required custom coding due to a lack of access to the optimized PyTorch NN libraries we used for the CNNs. As a result, the training speed of our LSTM network decreased even further, and a fair speed comparison was impossible.

\section{Additional value of the uncertainty in predictive process monitoring applications (RQ4)}\label{sec:working}
Knowledge of the uncertainty of the predictions is critical for practical predictive process monitoring applications. We identify three key drivers below.

\paragraph{Higher accuracy and acceptance of prediction systems}
Subsection~\ref{subsection:overall} revealed that our techniques generally improve the overall quality of the predictions. More importantly, Subsection~\ref{subsection:different} demonstrated that the uncertainties correlate strongly with the accuracy of the prediction. The latter result enables organizations to set arbitrary accuracy thresholds for their prediction systems. This can be achieved by computing the corresponding uncertainty threshold and retaining only those predictions that fall below this threshold. Predictions that fall short of the accuracy requirements are rejected and can be passed to humans or another system. Confidence intervals are an alternative method for countering the negative impact of uncertain predictions. In summary, by deploying our techniques, organizations can obtain better overall predictions and recognize and handle false, potentially dangerous predictions.

\paragraph{Improved human-machine symbiosis}
Knowledge of the uncertainty of the predictions enables two-track systems, in which cases with accurate predictions remain on the automated track, whereas cases with predictions that did not meet the uncertainty thresholds are moved to the human track. The latter cases are often more complex, rare and interesting. A two-track system could increase people's work satisfaction and better utilize their cognitive faculties.

\paragraph{Working with smaller datasets: earlier adoption of prediction systems}
Figs.~\ref{fig:uncertainty} and~\ref{fig:summary_results} show that insufficient data result in inaccurate predictions. As a result, organizations cannot use prediction systems until sufficient training data become available. This delay is detrimental in the current information age, which is characterized by rapid innovation, fast implementation and continuous learning. The determination of a certainty threshold and sorting predictions that do not meet this threshold allows prediction systems to be deployed more quickly. The initial proportion of accepted predictions will be minor but will gradually increase as the size of the dataset increases.
During this phase-in period, organizations can obtain invaluable information to improve their systems and data collection strategies.

\section{Conclusion and future work}\label{conclusions}
A BNN predicts a distribution. The mean of this distribution corresponds to the model's point estimate, while its variance corresponds to the prediction's epistemic uncertainty. BNNs are easy to implement because they are mathematically equivalent to deterministic NNs with dropout and L2 regularization. The heteroscedastic aleatoric uncertainty of the prediction can be determined by duplicating the last layer of the model and modifying its loss function (loss attenuation).

This paper demonstrated that in the field of predictive process monitoring, uncertainty estimates can successfully identify accurate and inaccurate model predictions. These results hold for both regression (remaining time prediction) and classification (outcome prediction) tasks. Early predictions of running cases are accompanied by uncertainty estimates, which is a crucial element of real-world cases. We also found that loss attenuation, dropout and the Bayesian approach all contribute toward higher-quality point estimates. Consequently, BNNs make better predictions than their deterministic counterparts. CNNs achieve better results on our datasets and tasks than LSTMs in a shorter amount of time. BNNs take approximately 40\% longer to train; however, their much smaller hyperparameter space compensates for this apparent disadvantage. Conventional, deterministic NNs can be improved with our techniques with minimal coding, and the resulting models perform similarly with less training data. Uncertainty allows users to set thresholds for retaining predictions that meet a required accuracy, build confidence intervals according to predictions, divide cases between computers and humans in intelligent ways, and adopt prediction models before large datasets have been collected. We believe that our results present compelling reasons to move predictive monitoring systems from academia to production. We hope that the uncertainty techniques presented in this paper will allow more data to be extracted from information systems in the near future.

Several topics in this emerging field merit further research. We used dropout to implement variational inference. However, other variational inference methods exist that may exhibit different characteristics. Furthermore, there are more applications uncertainty knowledge than those presented in this paper. For instance, uncertainty can be used to weigh loss function components in multitask neural network learning \cite{food} or multimodel ensembles. As we concentrated on the total uncertainty, evaluating the merits of the epistemic and aleatoric uncertainties is another path for future research. Ultimately, implementation in a production environment could pave the way for the widespread use of BNNs in predictive process monitoring.

\clearpage
\appendix
\setcounter{figure}{0}
\section{Early prediction bucketing}\label{app:bucket}
\begin{figure}[H]
\includegraphics[width=1\textwidth]{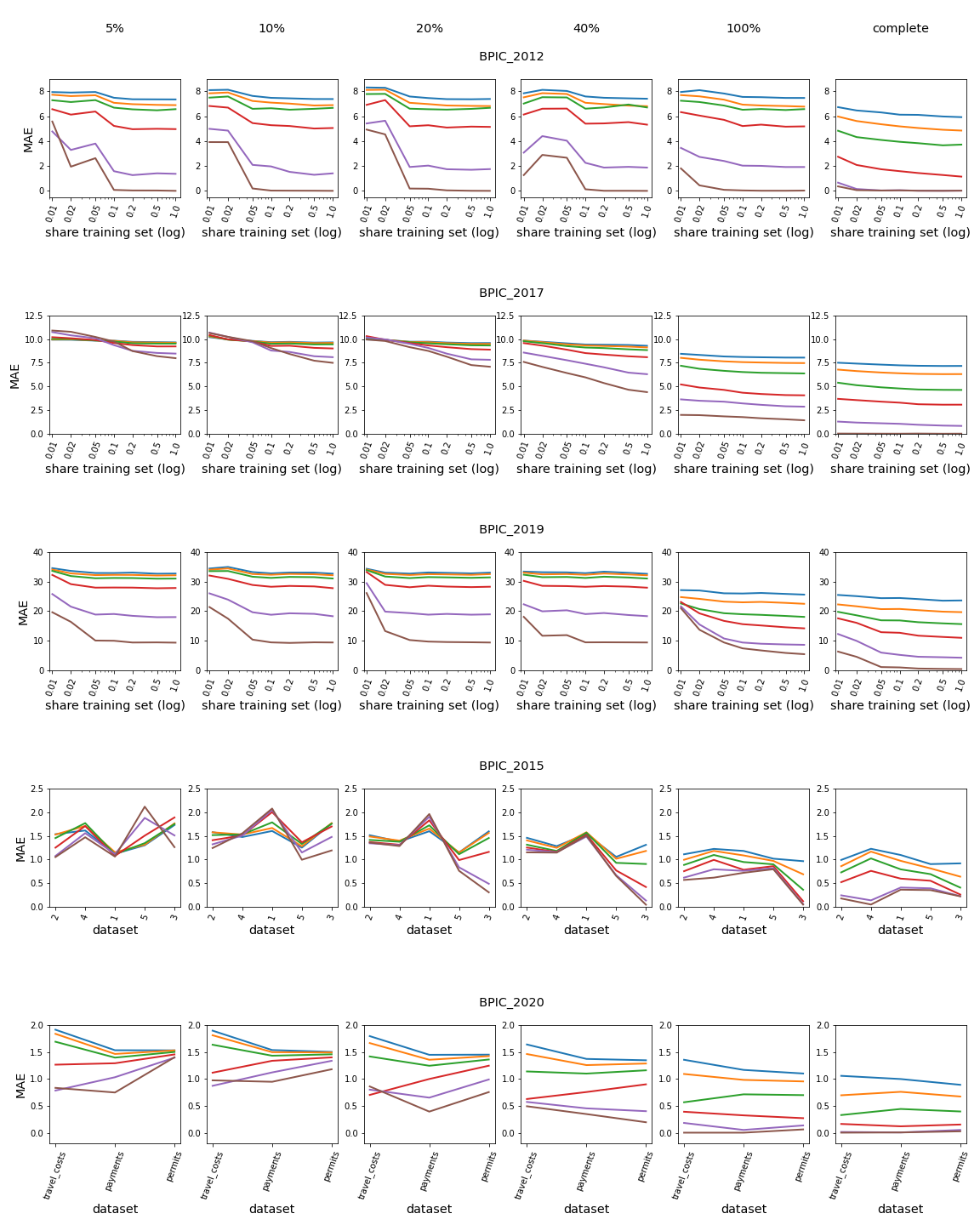}
\caption{Early predictions: same interpretation as Fig.~\ref{fig:earlycomplete}. The training was performed on buckets. The buckets consist of prefixes from the training that have shorter or equal durations to the longest prefix in each subgraph.} \label{fig:earlybuckets}
\end{figure}





\end{document}